\documentclass{sn-jnl}

\usepackage{graphicx}%
\usepackage{multirow}%
\usepackage{amsmath,amssymb,amsfonts}%
\usepackage{amsthm}%
\usepackage{mathrsfs}%
\usepackage[title]{appendix}%
\usepackage{xcolor}%
\usepackage{textcomp}%
\usepackage{manyfoot}%
\usepackage{booktabs}%
\usepackage{algorithm}%
\usepackage{algorithmicx}%
\usepackage{algpseudocode}%
\usepackage{listings}%
\usepackage{cite}

\usepackage{graphicx}
\usepackage{xcolor}
\usepackage{subcaption}
\usepackage{amsmath}
\usepackage{amsfonts}
\usepackage{algorithm}
\usepackage{algpseudocode}
\usepackage{tabularx}

\usepackage{tikz}
\usetikzlibrary{decorations.markings}
\usetikzlibrary{arrows}
\usetikzlibrary{decorations.pathreplacing}
\usetikzlibrary{shapes.geometric, arrows, positioning}
\tikzstyle{style} = [rectangle, minimum width=2cm, minimum height=0.6cm, text centered, draw=black]
\tikzstyle{style2} = [rectangle, minimum width=2cm, minimum height=0.6cm, text centered, draw=black, font=\bfseries]
\tikzstyle{arrow} = [thick,->,>=stealth]

\begin{document}

\title[Real-Time Motion Detection Using Dynamic Mode Decomposition]{Real-Time Motion Detection Using Dynamic Mode Decomposition}


\author[1]{\fnm{Marco} \sur{Mignacca}}\email{marco.mignacca@mail.mcgill.ca}

\author[2]{\fnm{Simone} \sur{Brugiapaglia}}\email{simone.brugiapaglia@concordia.ca}

\author*[2]{\fnm{Jason J.} \sur{Bramburger}}\email{jason.bramburger@concordia.ca}

\affil[1]{\orgdiv{Department of Mathematics and Statistics}, \orgname{McGill University}, \orgaddress{\city{Montr\'eal, QC}, \state{QC}, \country{Canada}}}

\affil[2]{\orgdiv{Department of Mathematics and Statistics}, \orgname{Concordia University}, \orgaddress{\city{Montr\'eal, QC}, \state{QC}, \country{Canada}}}


\abstract{Dynamic Mode Decomposition (DMD) is a numerical method that seeks to fit time-series data to a linear dynamical system. In doing so, DMD decomposes dynamic data into spatially coherent modes that evolve in time according to exponential growth/decay or with a fixed frequency of oscillation. A prolific application of DMD has been to video, where one interprets the high-dimensional pixel space evolving through time as the video plays. In this work, we propose a simple and interpretable motion detection algorithm for streaming video data rooted in DMD. Our method leverages the fact that there exists a correspondence between the evolution of important video features, such as foreground motion, and the eigenvalues of the matrix which results from applying DMD to segments of video. We apply the method to a database of test videos which emulate security footage under varying realistic conditions. Effectiveness is analyzed using receiver operating characteristic curves, while we use cross-validation to optimize the threshold parameter that identifies movement.}




\maketitle

\section{Introduction}

Dynamic Mode Decomposition (DMD) is a method for forecasting and extracting complicated spatio-temporal behaviour from high-dimensional datasets. First developed by Schmid \cite{schmid2010dynamic}, DMD seeks to identify a matrix $A$ so that advancing the data forward by one timestep is equivalent to multiplication by this matrix \cite{tu2014dmd,DMDbook}. In this way, DMD poses complex movement over space and through time as a linear discrete dynamical system via the matrix $A$. Since linear systems are entirely understood by their spectrum, the eigenvectors of $A$, called the {\em DMD modes}, can be seen to represent macroscopic patterns in the dataset that persist through time, either oscillating, growing, or decaying at a fixed rate determined by the associated eigenvalue. DMD has connections to the Koopman operator \cite{rowley2009spectral,brunton2022modern} and has proven to be a useful tool in identifying coherent structures in fluid mechanics \cite{schmid2011applications,menon2020dynamic,colbrook2023residual} and neuroscience \cite{brunton2016extracting,partamian2023analysis,shiraishi2020neural}, as well as to develop financial trading strategies \cite{mann2016dynamic}, among numerous other applications; see \cite{DMDbook,schmid2022dynamic} and the references therein. Improvements on the method include building in known physics \cite{baddoo2023physics,morandin2023port}, accounting for irregular timesteps and noisy measurements \cite{lee2023optimized,askham2018variable,hemati2017biasing}, and to incorporate the effect of control \cite{proctor2016dynamic,garcia2022extended}.    

\begin{figure}
     \centering
     \begin{subfigure}[b]{0.32\textwidth}
         \centering
         \caption{Original frame}
         \includegraphics[width=\textwidth]{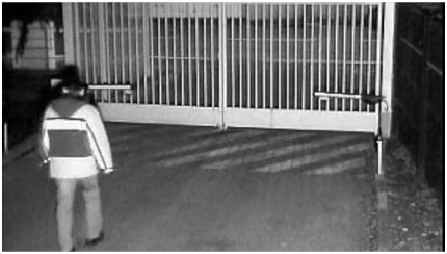}
         \label{fig:original_frame}
     \end{subfigure}
     \begin{subfigure}[b]{0.32\textwidth}
         \centering
         \caption{Background}
         \includegraphics[width=\textwidth]{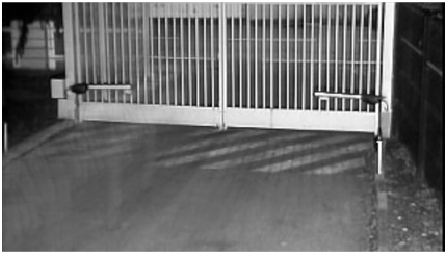}
         \label{fig:background_isolated}
     \end{subfigure}
     \begin{subfigure}[b]{0.32\textwidth}
         \centering
         \caption{Foreground}
         \includegraphics[width=\textwidth]{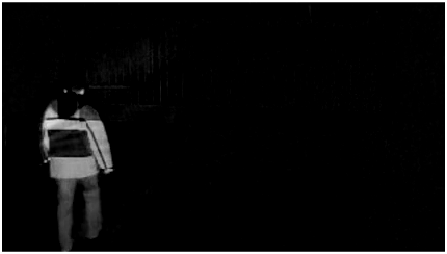}
         \label{fig:foreground_isolated}
     \end{subfigure}
        \caption{Screenshots of DMD being used to separate out background and foreground in video data. The video is taken from the Background Models Challenge database \cite{BMC}.}
        \label{fig:background_foreground}
\end{figure}

The utility of DMD has also made its way into computer vision. This includes employing DMD for edge detection in images \cite{bi2017dynamic}, salient region detection \cite{sikha2021dynamic,sikha2018salient,sikha2020multi}, and performing image classification \cite{rahul2019dynamic}. Most important to the work herein, DMD has been applied to separate out the background and foreground in videos, allowing for object tracking \cite{krake2022efficient,grosek2014dynamic,kutz2015multi,erichson2019compressed,haq2020dynamic,han2022quaternion}. Roughly, this comes from the fact that eigenvectors of the DMD matrix $A$ with eigenvalues having modulus near 1 represent patterns in the (vectorized) video data that exhibit little change from frame to frame. Thus, these eigenvectors can be seen as representing the background of the video data, while the remaining eigenvectors fill in the foreground motion. Importantly, projecting the video onto the background/foreground DMD modes constitutes a separation of these distinct components of the video. The result is then two videos: one of a nearly static background and another of just the isolated motion, as exemplified in Figure~\ref{fig:background_foreground}. 

\begin{figure*}[t]
    \centering
    \includegraphics[width=\textwidth]{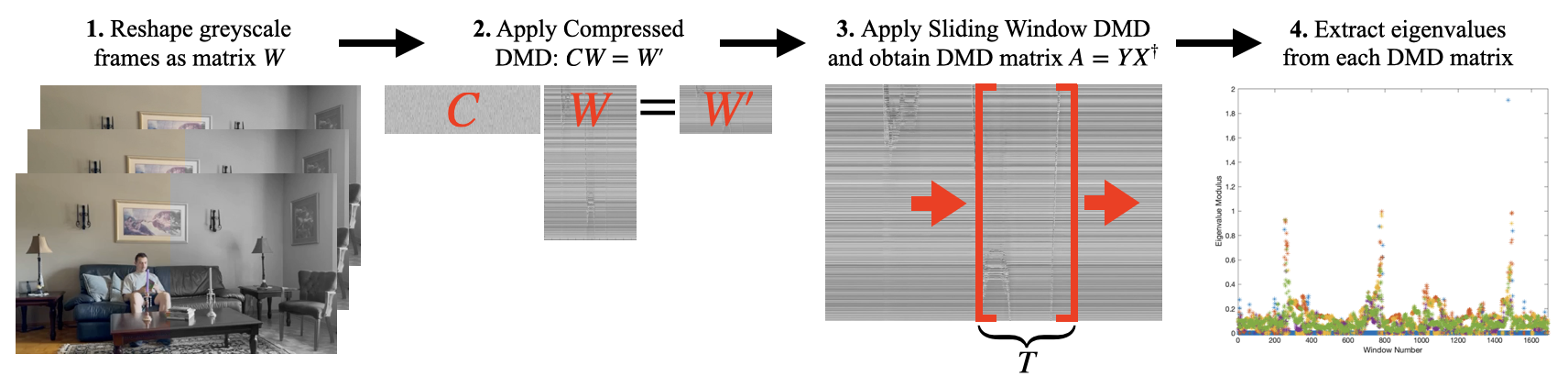}
    \caption{\color{black}A visual overview of real-time motion detection and tracking with DMD. Sequential collections of video frames, or windows, are fed in and compressed to reduce computational complexity. Then, DMD is applied to identify spatio-temporal features within the windows. Spikes in the spectrum of the DMD matrices over the windows indicate movement in the video, thus detecting events as they occur in real-time.}
    \label{fig:DMDalgo}
\end{figure*}

Our contribution in this paper is to present a method of real-time motion detection in streaming video data using DMD. To do this we apply DMD not to the entire video all at once, but to short windows as they are fed in. Precisely, we employ sliding window DMD \cite{dylewsky2019dynamic,bramburger2020sparse} to streaming video data to examine the dominant timescales in each segment of the video. These timescales are revealed by examining the modulus of the logarithm of the eigenvalues for the DMD matrix within each window. Sudden spikes in the DMD spectrum denote fast timescale movement in the foreground, allowing for motion detection. A visual overview of our method can be found in Figure~\ref{fig:DMDalgo}, while a proper explanation of the method can be found in Section~\ref{sec:DMDsurveillance} below. Once movement has been detected, one can then use the DMD matrix to isolate the foreground motion by subtracting out the background. Computation time is accelerated by compressing the frames of the video \cite{erichson2019compressed}, resulting in a computationally cheap and efficient method for real-time motion detection in streaming video data.  

Motion detection is a classic problem in computer vision with many complementary and competing methods. The simplest method identifies temporal differences \cite{cheung2005robust,yu2009real,kumar2024handling}, amounting to searching for significant change in the pixels from one frame to the next. This method is sensitive to abrupt changes in illumination, shadows, and repetitive motions, while also having the drawback that it cannot accurately extract the moving object boundaries. Alternatively, the work \cite{tsai2008motion} transforms the frames into Fourier space where object movement can be shown to create a non-vertical structure in the 2D Fourier transform. While again being an effective method, it suffers from a high computational cost. Computational cost can be brought down using the cosine transform \cite{oh2012real}, for which graph cuts can be employed to isolate movement in the video. Graph cuts come with their own well-documented limitations (see \cite{boykov2006graph} and the references therein), including significant memory usage. Another drawback of coupling motion detection algorithms with graph cuts is the simple fact that identifying and isolating movement becomes a two-step task. A major advantage of our method herein is that DMD provides the necessary tools to both identify movement (through eigenvalues) and isolate it (through eigenvectors). 

{\color{black} The review \cite{singh2020object} compares eight methods for real-time video surveillance of varying complexity. It is found that while simpler methods, such as examining temporal differences, are fast, they often fail to handle many of the complexities encountered in real-world video data. As expected, better performing methods come with higher computation time and a more complicated implementation. This includes a recent interest in applying neural networks to motion detection tasks \cite{fu2021camera,mandal2020motionrec,duong2023deep,liu2023amp}. This however comes at the cost of requiring a lot of training data to initialize the method and tedious hyperparameter tuning to obtain optimal results. This means that reproducing results is often difficult, while significant time and effort can be spent both curating training data and training the network itself. Even though we cannot claim that our DMD-based method herein can outperform state-of-the-art neural network or Gaussian mixture models \cite{bouwmans2008background,rakesh2023moving,nandhini2022cnn}, it offers simplicity and speed in its implementation, while also being backed up by dynamical systems theory to justify its performance. Fundamentally, DMD is a multiresolution analysis method that simultaneously accounts for spatial and temporal coherencies. It identifies macroscopic patterns in the data and is more robust to scale overlap than related methods \cite{dylewsky2019dynamic}, making it optimally suited for the task of motion detection. Furthermore, our array of test videos demonstrate that our method can be applied to situations that cause difficulties for others, such as multi-object tracking \cite{dai2022survey} and illumination variation.

}


The paper is organized as follows. In Section~\ref{sec:Methods} we provide an overview of DMD (Section~\ref{sec:DMD}, \ref{sec:cDMD}, \ref{sec:SlidingWindow}) and its capabilities for background subtraction (Section~\ref{sec:BackgroundSubtraction}), followed by a presentation of our own technique for motion detection in video (Section~\ref{sec:DMDsurveillance}). In Section~\ref{sec:Experiments}, we apply this method to a dataset of videos (Section~\ref{sec:Database}) and discuss its performance (Section~\ref{sec:PerformanceDemonstration}). The performance is then evaluated through the lens of receiver operating characteristic curves (Section~\ref{sec:ROC}), and a strategy to optimize a key parameter involving $k$-fold cross-validation is presented in Section~\ref{sec:ThresholdOptimization}. {\color{black} In Section~\ref{sec:OtherVideos} we showcase the performance of the method on a publicly available benchmark video database, and provide concluding remarks in Section~\ref{sec:Conclusion}.}

\section{Background on Dynamic Mode Decomposition}\label{sec:Methods}

In this section we provide an overview of all the relevant aspects of DMD for our method. We begin in Section~\ref{sec:DMD} with a presentation of DMD in its simplest form. We then turn to a review of DMD with compression in Section~\ref{sec:cDMD}, sliding window DMD in Section~\ref{sec:SlidingWindow}, and background subtraction in Section~\ref{sec:BackgroundSubtraction}. 

\subsection{Dynamic Mode Decomposition}\label{sec:DMD}

We begin with a review of DMD in its most basic form, while directing the reader to \cite{DMDbook} and the references therein for a more complete discussion. Assume that we have data snapshots $\{X_n\}_{n = 1}^{N+1} \subset \mathbb{C}^{M}$ for $N \geq 1$. In the coming applications these snapshots will be vectorized greyscale video frames, and so we make the assumption that they are arranged sequentially. Precisely, each snapshot $X_n$ is generated by the same process and sampled at times $t_n$ that are evenly spaced, i.e., $h := t_{n+1} - t_n > 0$ for all $n = 1,\dots, N$. DMD then seeks to identify a matrix $A \in \mathbb{C}^{M\times M}$ so that
\begin{equation}\label{DMDdynsyst}
	X_{n + 1} = AX_n, \qquad n = 1,\dots, N.
\end{equation}
That is, advancing from one snapshot to the next is equivalent to multiplying the current snapshot by the {\em DMD matrix} $A$. 

Identifying an appropriate DMD matrix is achieved by least squares regression. First, arrange the snapshots into $M \times N$ data matrices 
\begin{equation}\label{DMDsnapshots}
    \begin{split}
	   X &= \begin{bmatrix} X_1 & X_2 & \cdots & X_N \end{bmatrix}, \\ 
     Y &= \begin{bmatrix} X_2 & X_3 & \cdots & X_{N+1} \end{bmatrix}.
    \end{split}
\end{equation}
Notice that by definition each column of $Y$ is exactly one snapshot in the future from the snapshot in the corresponding column of $X$. Then, the DMD matrix $A$ is identified as the minimizer of
\begin{equation}
	\min_{A \in \mathbb{C}^{M \times M}} \|Y - AX\|^{2}_{F},
\end{equation}
where $\|\cdot\|_F$ denotes the Frobenius norm of a matrix. The minimizing DMD matrix is known to be $A = YX^\dagger$, where $X^\dagger$ is the Moore--Penrose pseudo-inverse of the data matrix $X$. Thus, upon arranging the snapshots into the matrices $X$ and $Y$, we can immediately obtain the DMD matrix through a quick and simple process of pseudo-inversion followed by matrix multiplication.  

An advantage of using DMD is that the resulting linear dynamical system \eqref{DMDdynsyst} can be completely understood in terms of the spectrum of the matrix $A$. Indeed, the correspondence \eqref{DMDdynsyst} between snapshots guarantees the eigenvector expansion
\begin{equation}\label{eq:eigen}
    X_{n} \approx \sum_{m=1}^{M} c_{m} \lambda_{m}^{n} \psi_{m},
\end{equation}
where $(\lambda_{m}, \psi_{m}) \in \mathbb{C} \times \mathbb{C}^{M}$ are the $M$ eigenpairs of $A$, and the coefficients $c_{m}$ are given by $\Psi^{-1}X_{1}$, where $\Psi = [ \psi_{1}\ \psi_{2} \ \dots \ \psi_{M} ]$. Expansion \eqref{eq:eigen} provides a space-time separation of variables for the data with eigenvectors describing macroscopic patterns, or coherent structures, that can only grow, decay, or oscillate in time according to their eigenvalue. Importantly, notice that eigenvectors with eigenvalues $|\lambda_m| \approx 1$ are modes that change little in time, leading to the interpretation that they are the `background' in the data. To better separate background and foreground modes we sort according to the `continuous-time' eigenvalues, given by
\begin{equation}\label{eq:ContEigs}
	\omega_m = \log(\lambda_m)/h, 
\end{equation}
where we recall that $h>0$ is the temporal spacing between consecutive snapshots. 
Thus, background modes have $|\omega_m| \approx 0$. Furthermore, the sum \eqref{eq:eigen} is equivalently written as 
\begin{equation}\label{eq:ContExpansion}
    X_{n} \approx \sum_{m=1}^{M} c_{m} \mathrm{e}^{\omega_m t} \psi_{m},    
\end{equation}
where $t = nh$.

\subsection{Compressed Dynamic Mode Decomposition}\label{sec:cDMD}

When snapshots are high-dimensional ($M \gg 1$), as in the case of high-resolution videos, applying DMD is computationally expensive due to the pseudo-inversion required to identify $A$. This computational bottleneck can be overcome using compressed DMD (cDMD) \cite{erichson2019compressed} which we incorporate into our method below. Begin by fixing an integer $p \ll M$ and generate a random measurement matrix $C \in \mathbb{C}^{p \times M}$. As explained in \cite{erichson2019compressed}, this matrix can be constructed componentwise via sampling a uniform or Gaussian distribution, among other choices (we employ the former, which is expressed as $C = \text{rand}(p,M)$ in MATLAB). With this matrix we compress our data matrices \eqref{DMDsnapshots} into $p\times N$ matrices
\begin{equation}\label{eq:compressed_data}
	X' = CX, \qquad Y' = CY.
\end{equation} 
To further eliminate redundancies in our data, we can rank-reduce the data using the Singular Value Decomposition (SVD) \cite{tu2014dmd}. This is achieved through the reduced SVD $X' = U_r\Sigma_rV^*_r$, where $U \in \mathbb{C}^{p \times r}$, $\Sigma \in \mathbb{C}^{r \times r}, V \in \mathbb{C}^{N \times r}$, with $r \ll N$ and $r \leq p$, and the subscript $r$ denotes only the first $r$ columns and/or rows of each matrix in the SVD of $X'$. This leads to an $r \times r$ DMD matrix given by 
\begin{equation}
    \tilde{A} = U^*_r Y' V_r \Sigma_r^{\dagger}.
\end{equation}
Notice that computation time is significantly reduced since $r \ll M$. 

Information from the compressed and rank-reduced matrix $\tilde{A}$ can be leveraged to obtain information about the original data in $X$ and $Y$. This comes from first letting $Q\in\mathbb{C}^{r\times r}$ be a matrix of eigenvectors of $\tilde{A}$ with the diagonal matrix of eigenvalues $\Lambda$ satisfying $\tilde{A}Q=Q\Lambda$. Full $M$-dimensional DMD eigenvectors are then recovered through the $M\times r$ matrix
\begin{equation}
    \Phi = YV\Sigma^{-1}Q,
\end{equation}
where we notably use the original data matrix $Y$. The columns of $\Phi$ contain the DMD modes needed to construct the space-time expansion \eqref{eq:eigen}, with corresponding eigenvalues on the diagonal of $\Lambda$. 

Overall, cDMD proves to be much more computationally efficient due a smaller SVD to calculate. The only additional computational costs comes from generating the measurement matrix $C$ and computing the respective matrix products $X'$ and $Y'$ as in (\ref{eq:compressed_data}). With our method of motion detection for video surveillance, generating $C$ comes as a one-time upfront cost that reduces numerous computations over time. This is because we will apply DMD to incoming windows of video data, each of which is compressed using the same compression matrix, thus reducing computation within each window. 

\subsection{Sliding Window DMD}\label{sec:SlidingWindow}

This idea of applying DMD in real-time to segments of video data is inspired by the sliding window DMD algorithm \cite{dylewsky2019dynamic}, which was originally devised to better resolve multiscale dynamics. In this framework, snapshots are separated into overlapping, consecutive chunks and DMD is applied on each. To be specific, the $N+1$ snapshots of the video are divided into windows of size $T >0$ so that each window is identified with a value $k \in \{1, \dots, N-T\}$. The $k$th data matrices have the form
\begin{equation}
    \begin{split}
	   X^{(k)} &= \begin{bmatrix} X_{k} & X_{k+1} & \cdots & X_{k+T} \end{bmatrix}, \\ 
     Y^{(k)} &= \begin{bmatrix} X_{k+1} & X_{k+2} & \cdots & X_{k+T+1} \end{bmatrix}.
    \end{split}
\end{equation}
With an eigenvector expansion as in equation (\ref{eq:ContExpansion}) in hand for each window, which we denote by $\bar{X}_{k}$ and which corresponds to the $k$th DMD matrix $A^{(k)} = Y^{(k)} (X^{(k)})^{\dagger}$, the entire video can be reconstructed via a linear combination of each windowed expansion weighted with Gaussians as follows:
\begin{equation}\label{eq:video_reconstruct}
    X_{\text{full}} = \frac{\sum_{k=1}^{N-T} e^{-(t - \eta_{k})/\sigma^{2}}\bar{X}_{k}}{\sum_{k=1}^{N-T}e^{-(t - \eta_{k})/\sigma^{2}}},
\end{equation}
where $\eta_{k}$ is the midpoint of the $k$th window, $t$ is a vector consisting of the temporal grid, and the standard deviation $\sigma$ of the Gaussian is prescribed. Since each iteration of this method only requires information from one of the chunks of video, this can easily be generalized to real time streaming data. A pseudocode of the DMD procedure, combining cDMD and sliding window DMD, is presented in Algorithm~\ref{alg:DMD}.

\subsection{Background Subtraction with DMD}\label{sec:BackgroundSubtraction}

An impressively successful application of DMD is background separation in video data \cite{han2022quaternion,kutz2015multi,grosek2014dynamic,krake2022efficient}. The method takes advantage of the linearity of DMD by separating out the fast- and slow-moving modes in \eqref{eq:eigen}. Recalling that the eigenvectors of the DMD matrix correspond to coherent structures in the data, eigenvalues with modulus near 1 are associated with components that exhibit little change over time. In the context of a video filmed with a fixed camera position, the portion of the video with unchanging dynamics constitutes the background, while the remaining eigenvectors are associated with the foreground. To this end, we are able to approximate the background by considering \eqref{eq:eigen} summed over only eigenvectors with eigenvalues near 1, sometimes referred to as the {\em low-rank component} of the data. Summing over the foreground constitutes the {\em sparse component}.

\begin{figure}
    \centering
    \includegraphics[width=0.6\textwidth]{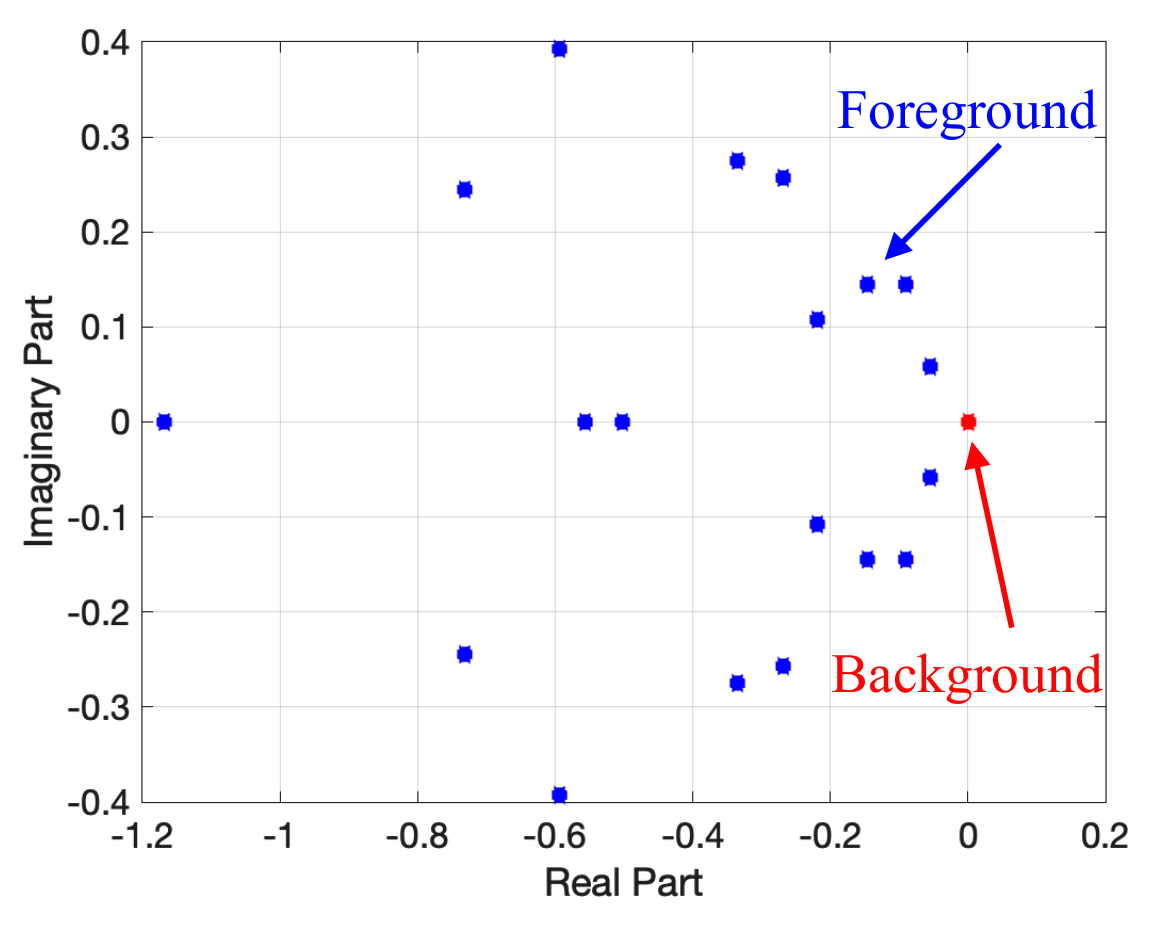}
    \caption{Plot of the continuous-time eigenvalues \eqref{eq:ContEigs} in the complex plane (corresponding to the 100th window) for the video in Figure~\ref{fig:background_foreground} with $r=20$. The single eigenvalue in red close to the origin constitutes the background mode, while the remaining 19 eigenvalues in blue make up the foreground motion.}
    \label{fig:continuous_eigenvalues}
\end{figure}

To properly separate out low-rank and sparse components from video data we perform the following. First, define a threshold $\varepsilon > 0$ that separates out fast and slow frequencies \eqref{eq:ContEigs}. We set 
\begin{equation}
    \tilde{L} = \sum_{k:\hspace{0.1cm} \vert \omega_{k} \vert \leq \varepsilon} c_{k} \psi_{k} e^{\omega_{k}t},
\end{equation}
and note that with real video data it is often the case that the sum defining $\tilde{L}$ is over only a single eigenvector. This comes from a clear distinction between frequencies with $\omega_k \approx 0$ and those whose frequencies are bounded away from zero, providing robustness with respect to the choice of $\varepsilon$. 
The reader is referred to Figure \ref{fig:continuous_eigenvalues} for a demonstration of this. Following \cite{grosek2014dynamic}, to ensure that the sparse component is real-valued, we set 
\begin{equation}
    \tilde{S} = X_n - |\tilde{L}|,
\end{equation}
where $X_n$ is the $n$th video frame and $|\tilde{L}|$ is the element-wise modulus of $L$. However, this process may result in negative pixel values in $\tilde{S}$ and so we define the residual $R$ to be a matrix of all such negative values in $\tilde{S}$. This leads to the true background $L$ and foreground $S$ components of the video, defined as
\begin{equation}\label{eq:LandS}
    L = R + |\tilde{L}|, \qquad S = \tilde{S} - R.  
\end{equation}
We empirically observed that defining $L = |\tilde{L}|$ resulted in improved background separation compared to defining $L$ as in (\ref{eq:LandS}), and so we used this formulation. This method works well, as evidenced by Figure~\ref{fig:background_foreground}. In practice, this procedure is applied in tandem with sliding window DMD on the reconstruction as in (\ref{eq:video_reconstruct}). In this instance however, we only sum over the slow and fast frequencies in the eigenvalue/eigenvector expansions (\ref{eq:ContExpansion}) to reconstruct the background and foreground, respectively.

\section{Motion Detection with DMD}\label{sec:DMDsurveillance}

Having reviewed the necessary DMD-related methods, we now present our method of motion detection. This application of DMD is the main contribution of this paper and the validation of the method is left to the next section. Recall that a visual overview of the method was presented in Figure~\ref{fig:DMDalgo}. 

\begin{algorithm}[t]
    \caption{Windowed cDMD}\label{alg:DMD}
    \textbf{Input:} $W \in \mathbb{R}^{M \times (N+1)}$, $r$, $p$, $T$ \\
    \ \textbf{Output:} $\Omega$ 
    \begin{algorithmic}
    \State $\text{num\_windows} \gets N + 1 - T$
    \State $C \gets \text{rand}(p, M)$ 
    \State $W' \gets C W$
    \State $\Omega \gets \text{zeros}(r, \text{num\_windows})$
    \For {$j = 1, \hdots, \text{num\_windows}$}
        \State $Y_{1} \gets W'(:, j:j+T-1)$
        \State $Y_{2} \gets W'(:, j+1:j+T)$
        \State $[U, \; \Sigma, \; V] \gets \text{svd}(Y_{1})$
        \State $[U_{r},\; \Sigma_{r}, \;V_{r}] \gets [U(:,1:r),\; \Sigma(1:r, 1:r),\; V(:,1:r)]$
        \State $\tilde{A} \gets U_{r}^{T} Y_{2} V_{r} \Sigma^{-1}$
        \State $\lambda \gets \text{eigenvalues}(\tilde{A})$
        \State $\omega \gets \vert \log{\lambda} \vert$
        \State $\Omega(:,j) \gets  \omega $
    \EndFor
    \end{algorithmic}
\end{algorithm}

The method starts by applying cDMD to windowed subsets of the streaming video data, which has been transformed into greyscale and vectorized into a matrix $W \in \mathbb{R}^{M \times (N+1)}$. The compression matrix $C$ and target rank $r$ are held constant over all windows, and each window is comprised of $T > 0$ consecutive frames of video. Algorithm~\ref{alg:DMD} provides our windowed cDMD method where we see that the output is the modulus of the continuous-time eigenvalues \eqref{eq:ContEigs} within each window. The scope of this work is primarily limited to video data where the camera position remains fixed, meaning the background is assumed to be relatively unchanged over the frames. This means that if there is little or no motion in the video, the eigenvalues within the window will all correspond approximately to the background, i.e., $\omega_i \approx 0$. However, if motion is present in a window one expects to see this represented as at least one of the eigenvalues deviating from the background cluster. As one sweeps across windows, motion entering the frames is expected to be represented by a large spike in the DMD matrix spectrum, as shown in Figure~\ref{fig:detection} below.

Motion detection from the DMD spectrum is automated by considering the sets of eigenvalues $\{\omega_i^{(k)}\}_{i = 1}^r$ and $\{\omega_i^{(k+1)}\}_{i = 1}^r$ from the consecutive $k$th and $(k+1)$th windows. We compute the averages
\begin{equation}\label{eq:AvgEigs}
    a_k = \frac{1}{r} \sum_{i = 1}^r |\omega_i^{(k)}|, \qquad a_{k+1} = \frac{1}{r} \sum_{i = 1}^r |\omega_i^{(k+1)}|.
\end{equation}
Then, defining a threshold $\Delta^* > 0$, motion is flagged in the $(k+1)$th window if
\begin{equation}\label{eq:RelativeChange}
    \bigg|\frac{a_{k+1} - a_k}{a_k}\bigg| \geq \Delta^*.
\end{equation}
By tracking the relative change in the mean of the eigenvalues over windows we allow for the method to exclude background motion that persists over frames, such as leaves being rustled by the wind.  
Note that we need only save information from the $k$th and $(k+1)$th consecutive windows at a time, while if no motion is detected we discard information from the $k$th window and compare the spectrum of the $(k+1)$th and $(k+2)$th windows. Combined with the data reductions from cDMD, this means that only two $r \times r$ matrices are stored in memory to perform the method. 

{\color{black}In practice, it was observed that foreground motion in video was sometimes accompanied by multiple consecutive activations; that is, the moduli of the eigenvalues would continue to increase in a few of the windows immediately following that of the initial detection. We believe that this is related to slow motion in the video, but we do not have a clear explanation for it. } One way this can be resolved is by storing the (binary) results of the motion detection test for the previous few windows and additionally checking that no motion was already flagged in these previous windows. 

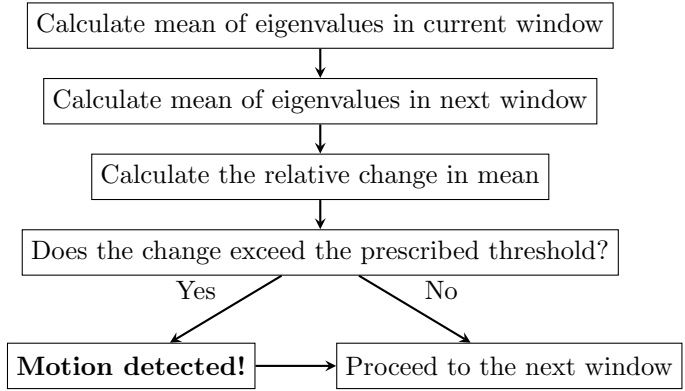
\begin{figure}[ht]
    \centering
    \begin{tikzpicture}[scale=0.5]
        \node (start) [style] {Calculate mean of eigenvalues in current window};
        \node (framek1) [style, below of=start] {Calculate mean of eigenvalues in next window};
        \node (compare) [style, below of=framek1] {Calculate the relative change in mean};
        \node (decision) [style, below of=compare] {Does the change exceed the prescribed threshold?};
        \node (motion) [style2, below of=decision, xshift = -2.5cm, yshift=-0.5cm] {Motion detected!};
        \node (framek2) [style, below of=decision, xshift=2.5cm, yshift = -0.5cm] {Proceed to the next window};

        \draw [arrow] (start) -- (framek1);
        \draw [arrow] (framek1) -- (compare);
        \draw [arrow] (compare) -- (decision);
        \draw [arrow] (decision) -- node[anchor=south east] {Yes} (motion);
        \draw [arrow] (decision) -- node[anchor=south west] {No} (framek2);
        \draw [arrow] (motion) -- (framek2);
    \end{tikzpicture}
    \caption{Flowchart demonstrating the motion detection scheme.}
    \label{fig:flowchart}
\end{figure}

Finally, if motion is detected in a window, we apply the method of background separation from Section~\ref{sec:BackgroundSubtraction}. This allows for the isolation of the detected motion, again using only information from the stored $r\times r$ DMD matrix corresponding to the window where motion is first detected. This is summarized in the flowchart in Figure~\ref{fig:flowchart}. 

This method involves initializing a number of parameters that may affect its performance. Specifically, the target rank $r$ dictates the number of eigenvalues in each window, while the target dimension $p$ from cDMD, the window length $T$, and most importantly, the detection threshold $\Delta^*$ have an impact on performance. The choice of parameters that leads to optimal performance of the method is problem-dependent and should be optimized with simple tests. This might include designing motion to be captured by the method, such as having someone walk into the frame, while also decreasing the sensitivity to exclude frivolous motion that might be unique to the application. In the following section we demonstrate the method's performance on real video data, while also evaluating performance with receiver operating characteristic curves and tuning the threshold detection parameter with $k$-fold cross-validation.

\section{Experimental Validation}
\label{sec:Experiments}
In this section we demonstrate the performance of our method on real video data. A full code repository is available to reproduce the results of this section. This includes all files used for our tests, as well as full MATLAB scripts to perform the method on video data and all the following validations: \href{https://github.com/marco-mig/dmd-motion-detection}{https://github.com/marco-mig/dmd-motion-detection}. 

In what follows, we fix the parameters $T$, $p$, and $r$, while focusing on the importance of the threshold parameter $\Delta^*$. Specifically, we take $p = 20$ and $r = 5$, which means that each DMD matrix is only $5 \times 5$. Thus, eigenvalues within each window can be computed quickly without presenting a bottleneck to the real-time nature of the algorithm. The window sizes are fixed to be $T = 80$, corresponding to 8/3 seconds. The random matrix $C$ is constructed only once and applied to all videos for consistency.

\subsection{Dataset}
\label{sec:Database}

\begin{figure}
    \centering
    \begin{subfigure}{0.45\textwidth}
        \centering
        \includegraphics[width=\linewidth]{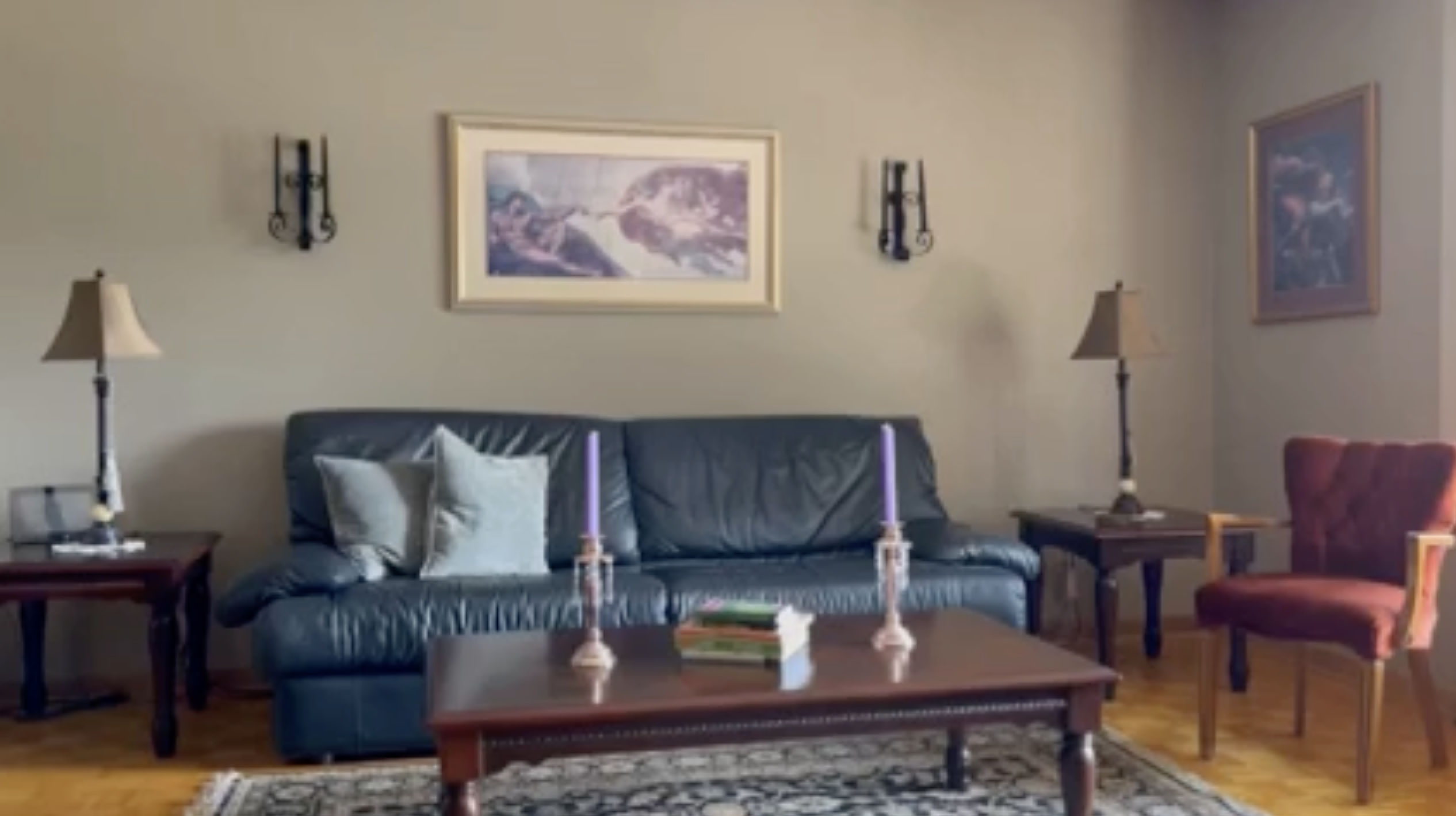}

    \end{subfigure}
    \begin{subfigure}{0.45\textwidth}
        \centering
        \includegraphics[width=\linewidth]{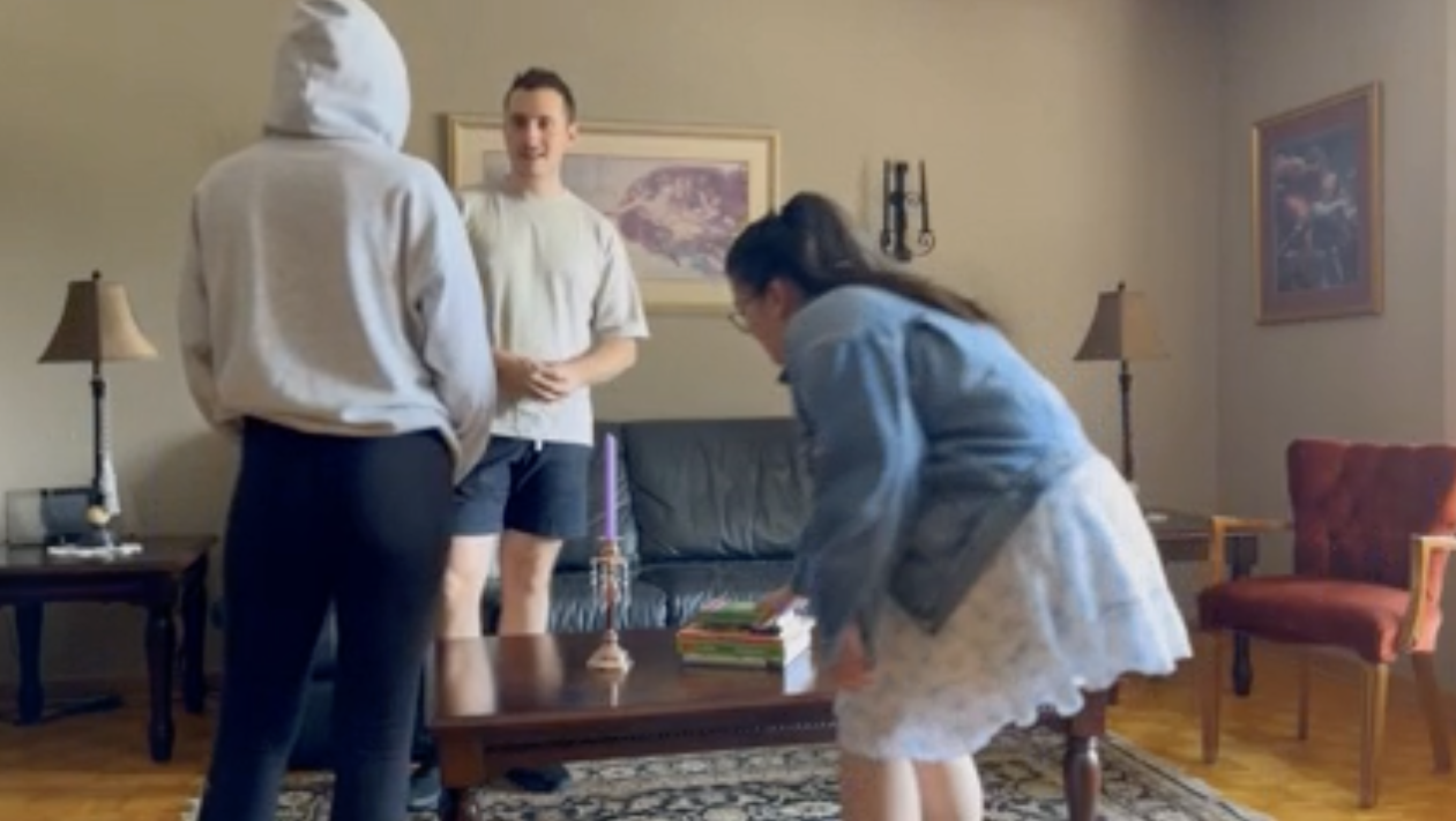}
    
    \end{subfigure}
    \\
    \begin{subfigure}{0.45\textwidth}
        \centering
        \includegraphics[width=\linewidth]{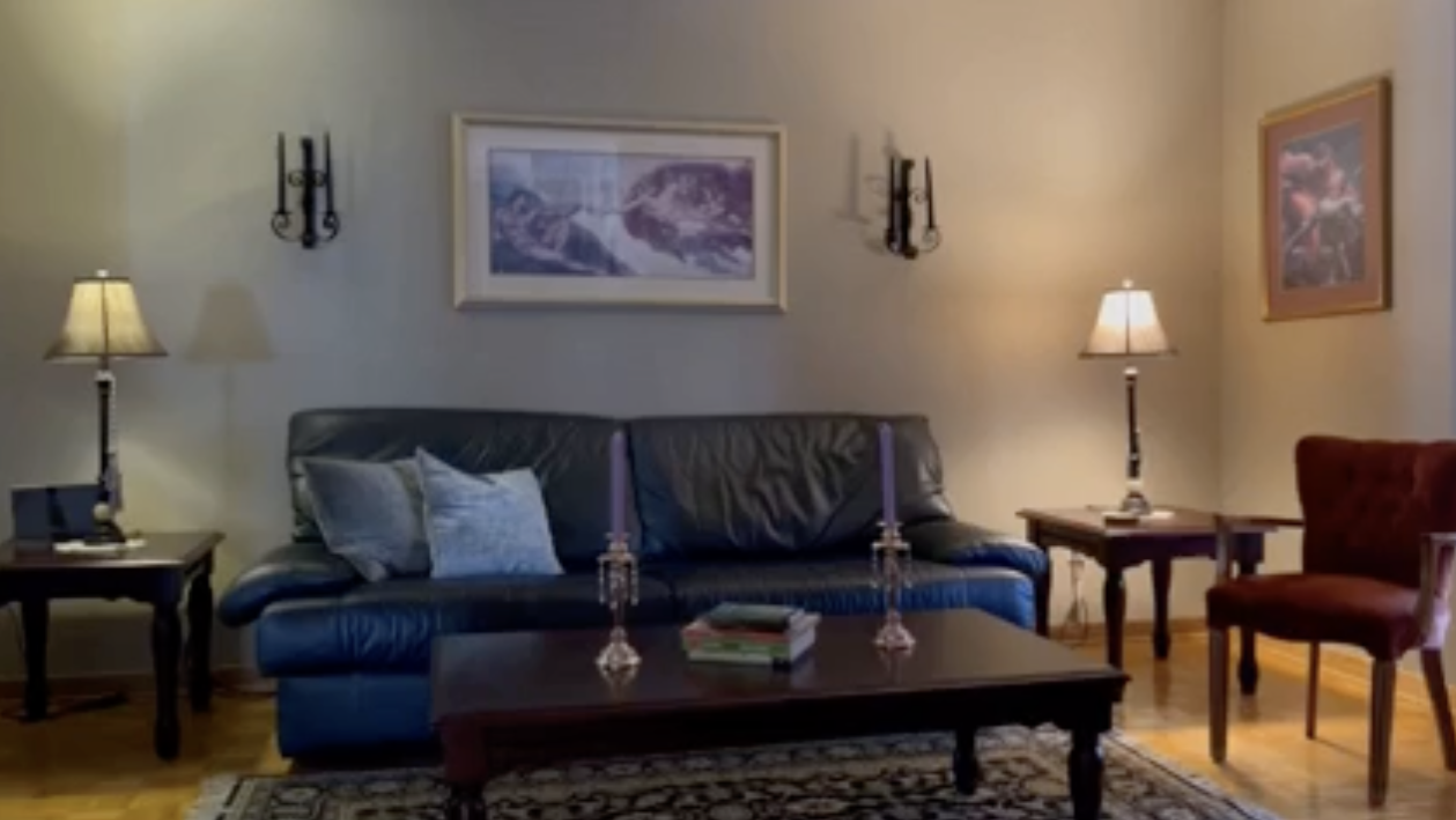}

    \end{subfigure}
    \begin{subfigure}{0.45\textwidth}
        \centering
        \includegraphics[width=\linewidth]{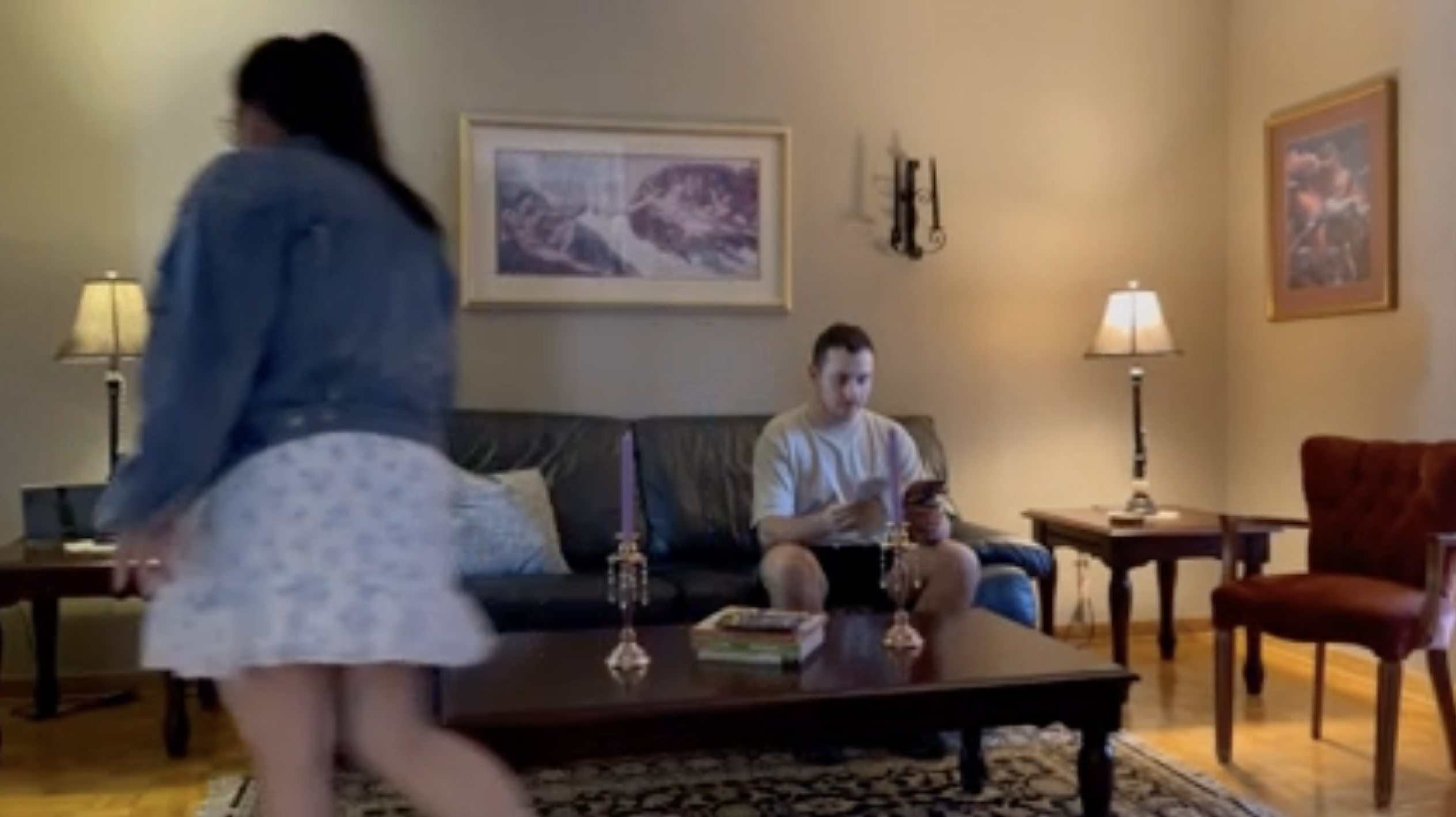}
    \end{subfigure}
    \caption{Sample frames of videos in database. Top: Background (left) and motion (right) in a well-lit video. Bottom: Background (left) and motion (right) with indoor lighting producing shadows.}
  \label{fig:videos}
\end{figure}

Initial tests of the method featured free videos from the Background Models Challenge database \cite{BMC}, including the one featured in Figure~\ref{fig:background_foreground}. However, to ensure a diversity of situations across the test set, we have opted to film our own videos using the built-in camera on the iPhone 13. Our videos were filmed to simulate security footage in that the camera is held still and various individuals enter and exit the frame to be detected. These videos include numerous variations such as the speed of entry and/or exit of the individuals, the distance of the motion from the camera, the time the individuals spend in the frame, and the actions performed by the individuals in the frame. Our test set includes a total of 20 videos, of which 15 were filmed in daylight and 5 with indoor lighting. Videos are of size 426 by 240 pixels and range in length from 27 seconds (810 frames) to 109 seconds (3270 frames). Each video begins and ends with a few seconds of a still background shot to provide a baseline for the method to gauge performance. Frames from videos in two lighting conditions are provided in Figure~\ref{fig:videos}.

{\color{black} We use this video database to gauge the performance of our method and discuss optimal parameter tuning. In Section~\ref{sec:OtherVideos} we will then turn to the Microsoft Wallflower dataset \cite{toyama1999wallflower} to provide further performance tests on a benchmark data set. We emphasize that for each new setting or camera placement one will be required to re-tune the parameters since the behavior of our method of motion detection depends on the specifics of the surrounding area. This means that one should not apply this method on streaming video footage without first optimally tuning the parameters on test footage from the same camera. This will be more apparent in Section~\ref{sec:OtherVideos} and, in particular, Table~\ref{tab:microsoft_wallflower}, where one sees that different threshold parameters are required to detect motion in the Microsoft Wallflower dataset due to the specific characteristics of each video, such as latent motion in the background.}

\begin{figure}[t]
    \centering
    \includegraphics[width=0.9\textwidth]{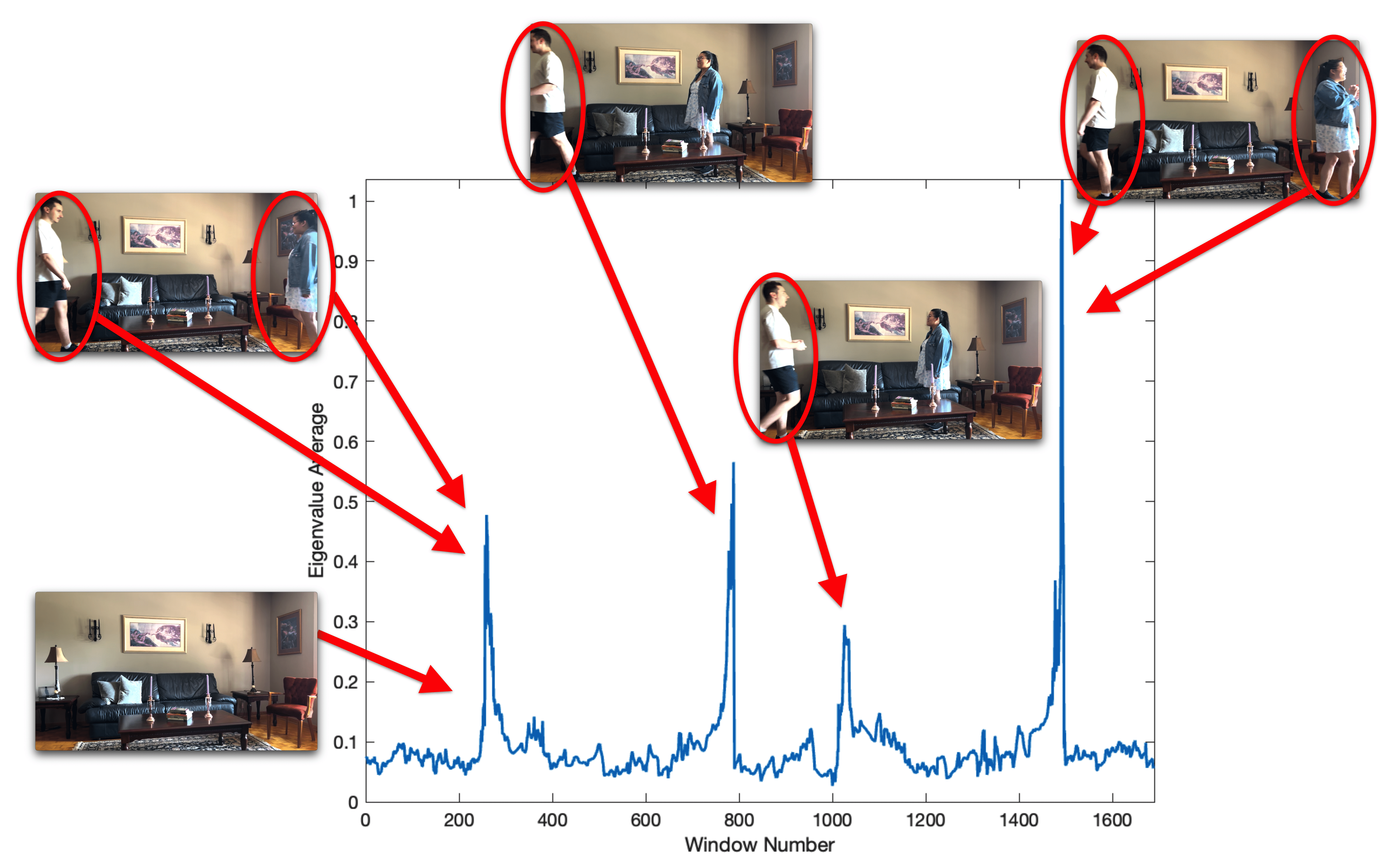}
    \caption{Sudden spikes in the average of the DMD spectrum represent motion detection. Each spike is accompanied by one or more individuals entering or exiting the frame, as demonstrated by the screenshots. }
    \label{fig:detection}
\end{figure}

\subsection{Performance Demonstration}
\label{sec:PerformanceDemonstration}

Our experiments demonstrate that sudden motion from an individual entering or exiting the video frame is accompanied by a spike in the average eigenvalues, as measured by \eqref{eq:RelativeChange}. A characteristic demonstration is provided in Figure~\ref{fig:detection} where one can clearly see motion detection through large deviations in the DMD spectrum. {\color{black} When an individual enters the frame, it is reflected in an eigenvalue spike corresponding to the frame number minus the window length $T$. This is when the frame at which the individual enters the video first appears in the window as the rightmost column of the data matrix $W$, using the notation of Algorithm~\ref{alg:DMD}. Alternatively, when an individual exits the video, the window number where the eigenvalue spike takes place corresponds to the frame number itself. This is because it is the last window in which information about motion is present in the data matrix $W$, and so the change occurs after this information (in the left column of the data matrix) vanishes in the subsequent window. In summary, there is a temporal delay of length $T$, the window size, in detecting when an object has left the frame. Thus, a distinction is made between objects entering and exiting the frame, and the frame numbers supplied to the algorithm that calculates the error score in the following subsection have been adjusted accordingly.}

\subsection{Performance Evaluation}
\label{sec:ROC}
The task of identifying motion in windowed subsets of the video can be interpreted as a binary classification problem, since there are two distinct classification outcomes: either the window contains motion, or it does not. In what follows, we refer to the event of motion being present in a window as a positive event, and the lack thereof as a negative event. To this end, there are two possible errors that can be observed: either the algorithm classifies a negative event as positive, a false
positive (FP), or it classifies a positive event as negative, a false negative (FN). We also denote a successfully classified positive as a true positive (TP) and a successfully classified negative as a true negative (TN). 

To display the relationship between the true positive rate (TPR) and false positive rate (FPR) we will use a receiver operating characteristic (ROC) curve (see, e.g., \cite{stat}). The TPR can be thought of the rate at which the algorithm successfully classifies positive events as positive, while the FPR is the rate at which the algorithm classifies negative events as positive. A perfect classifier necessarily has a TPR of 1 and an FPR of 0, and the ROC curve for such a classifier is the graph of the indicator function of the interval $(0,1]$. Our ROC curve is traced out by recording the TPR and FPR of the algorithm over threshold values $\Delta^*$ ranging from $10^{-10}$ to $10^{10}$. In this section, we do not implement the fix for consecutive activations as described in Section~\ref{sec:DMDsurveillance} as this interferes with the expected behaviour of the algorithm at threshold values close to zero. That is, for these values, activations at nearly every (if not every) window containing motion are expected. Finally, a half-second tolerance (forward and backward) in detecting each event is imposed. 

\begin{figure}[t]
    \centering
    \includegraphics[width = 0.5\textwidth]{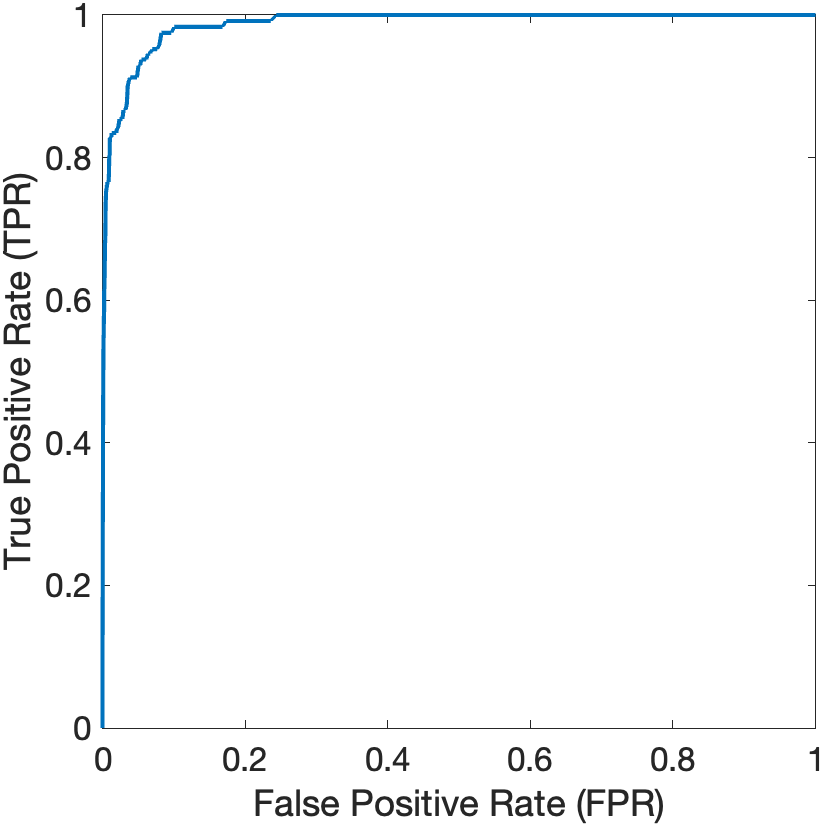}
    \caption{Average of the ROC curves generated from the 20 videos in the database.} 
    \label{fig:ROC}
\end{figure}

We calculated ROC curves for all 20 of the videos in our database using our motion detection method. The mean of all the ROC curves is presented in Figure~\ref{fig:ROC}. The performance of the classifier is quantified by the area under the ROC curve, which for the averaged ROC curve in Figure~\ref{fig:ROC} is $0.9876$. Since this value is quite close to the optimal performance at 1 and far from the random chance performance at 0.5, we find strong evidence for the quality of our method for detecting motion within windowed subsets of video data.

\subsection{Optimizing the Threshold}
\label{sec:ThresholdOptimization}

While our method contains a number of parameters to be set by the user, none is as important as the threshold $\Delta^*$. In this subsection we present a method for determining an optimal threshold to get the best performance from the method using a database of test videos that should be curated by the user before applying the method for video surveillance. We begin by defining an appropriate error measure which can then be minimized by the optimal threshold.

\subsubsection{Measuring Error}
\label{sec:MeasuringError}
Recall from Section~\ref{sec:ROC} that there are a total of four possible outcomes of the algorithm:
either it classifies correctly (encompassing TPs and TNs) or incorrectly (FPs and FNs). In many applications, these two latter outcomes are not equally costly. This is because in the context of video
surveillance, a FN, such as not detecting an intruder, can be much more costly than a FP, such as falsely indicating the presence of an intruder. Thus, our goal here is to present a possible error metric that takes this into consideration.

We begin by fixing a parameter $d^* > 0$. To obtain the error, we run the method over all videos in the database and check if the detection system has flagged motion in a window within plus or minus $d^*$ windows of where motion is known to have taken place. Thus, $d^*$ functions as the tolerance in lag for which the user allows the algorithm to detect motion.  The result is a tally of the total number of errors (FP + FN), for which we define the error score as
\begin{equation}\label{Error}
    E = \mathrm{FP} + c\cdot \mathrm{FN}.
\end{equation}
Here $c > 0$ is a weight parameter. Per the discussion in the previous paragraph, one typically would like $c > 1$, and often $c \gg 1$, to reflect the fact that a FN is more costly than a FP in a real-world situation. Hence, for a given threshold $\Delta^*$ the error $E$ is directly dependent upon both $d^*$ and $c$.

\subsubsection{A Variant of $k$-Fold Cross-Validation}
\label{sec:CrossValidation}
We will use a variant of $k$-fold cross-validation (see, e.g., \cite{stat}) to determine an optimal threshold value using our error metric \eqref{Error} and database of videos. In what follows we fix $d^* = 30$, which corresponds to a tolerance of detection within 1 second of an event taking place. The penalty ratio in the error metric is taken to be $c = 100$. 

Recall that our goal is to optimize the threshold $\Delta^*$, rather than estimate the error of the method applied to unseen data (which is the standard objective of cross-validation). However, these two objectives are related, as the error of
our motion detection algorithm will inform the choice of threshold. Thus, we modify the framework of $k$-fold cross-validation to incorporate some minimization steps to function as an algorithm to solve for an optimal threshold $\Delta^*$. Our implementation is as follows: 
\begin{enumerate}
    \item Shuffle the dataset randomly, and split it into $k$ folds (i.e., groups) of roughly equal size. Also, initialize a vector $P$ containing the threshold values to test over.
    \item Then, for each iteration of $i =1, \dots, k$: 
    \begin{enumerate}
        \item Select one fold as the validation set, and group the $k-1$ others into the training set. 
        \item For each video in the training set, compute its error score for each threshold value $\Delta \in P$. 
        \item Take the mean of the error scores for each threshold value tested, and find the threshold value $\Delta_{i}$ corresponding to the smallest average error. 
        \item Apply this threshold value to the videos in the validation set, and calculate the average error score over all these videos, denoted $E_{i}$.
    \end{enumerate}
    \item The result of this procedure is an array of threshold values $\Delta_{i}, \; i=1, \dots k$ and their corresponding errors $E_{i}$. The optimal threshold $\Delta^{*}$ is selected as the $\Delta_{i}$ corresponding to the smallest average error (from testing on the validation set). 
\end{enumerate}

\begin{table}[t]
    \centering
    \begin{tabularx}{0.7\textwidth}{l|XXXX}
        \toprule
        Iteration ($i$)& 1 & 2 & 3 & 4 \\
        \midrule
        Threshold $\Delta_i$ & 0.196  &  0.129  &  0.19  &  0.342 \\
        Error $E_{i}$ & 67.8 & 72.2 &  49.0 & 170.8 \\
        \bottomrule
    \end{tabularx}
    \caption{Results of our $4$-fold cross-validation for the threshold $\Delta_i$ at each iteration on the database.}
    \label{tbl:Results}
\end{table}



We implement the above with $k = 4$ since our dataset is relatively small, and we initialize $P$ with 1000 evenly-spaced values between 0 and 1. Table~\ref{tbl:Results} presents our results for one application of the modified $k$-fold cross-validation procedure. It can be seen that the optimal threshold corresponds to an error score of 49, where we recall that FP and FN have a score of 1 and 100, respectively. However, due to the small size of the video databases, there are large discrepancies between the error scores corresponding to
the optimal thresholds. Moreover, each possible arrangement of the database into the training and validation sets results in different values. We expect that this cross-validation method will produce a more representative set of optimal thresholds and corresponding errors when applied to a larger database.

\begin{figure}[t]
     \centering
     \includegraphics[width=0.5\textwidth]{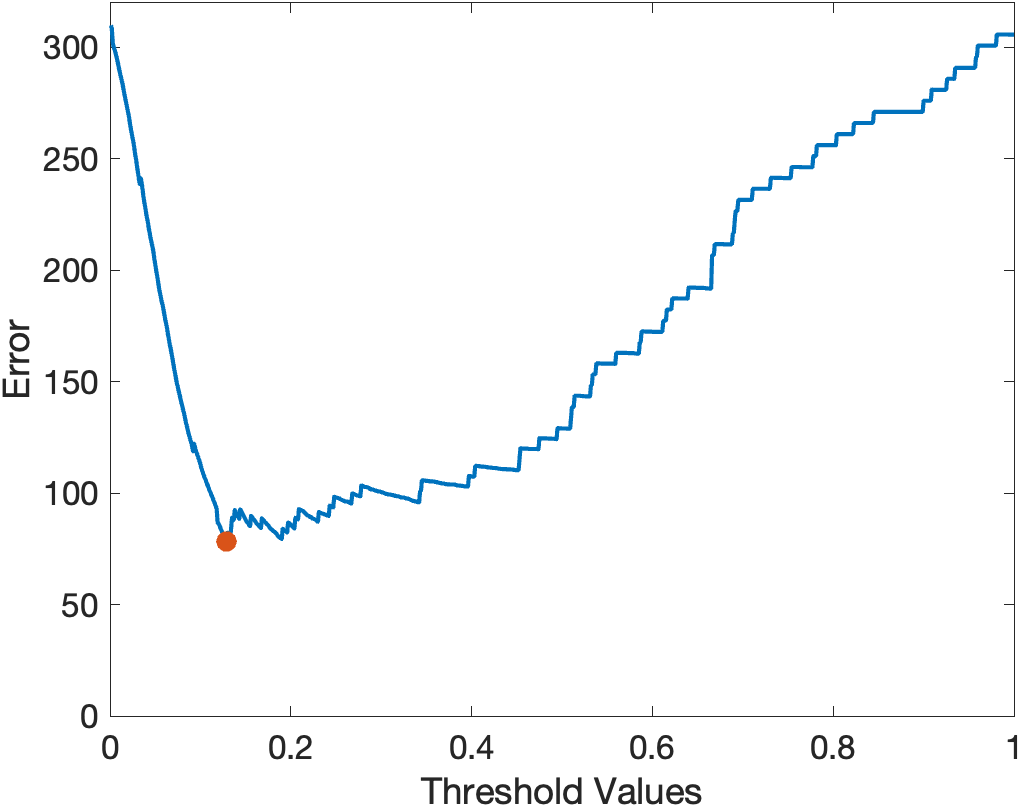}
     \caption{The average error on the training set over all folds as a function of the threshold value for the database. The optimal threshold value is indicated by a red dot. 
     }
   \label{fig:vcurves}
\end{figure}

From these results, two observations can be made. First, while the eigenvalue detection system can be effective when optimized for a single video, as shown in Figure~\ref{fig:detection}, when optimizing over a set of videos, the
performance tends to decrease. In particular, more FPs are made in order to minimize the occurrence of FNs over all the videos in the training set. This is evidently a consequence of the weight given to FN errors, and is thus dependent on the application. The second observation is that the optimal thresholds and their corresponding errors are always a trade-off between FPs and FNs, since they vary heavily depending on how each type of error
is weighed according to the value of $c$. As demonstrated in Figure~\ref{fig:vcurves}, plotting the average error over each of the training sets to which the cross-validation procedure is applied results in a curve with a distinct global minimum. Large error is observed for very small and very large threshold values due to the abundance of FP and FN errors, respectively.

\subsection{Application to Benchmark Video Database}\label{sec:OtherVideos}

{\color{black}
The previous demonstrations were carried out on a carefully curated database of videos that sought to probe performance based on variations in lighting, movement speed, and the number of objects moving. In this subsection we further demonstrate the performance of the method by implementing it on a benchmark set of publicly available videos. We use the Microsoft Wallflower dataset \cite{toyama1999wallflower}, which was recently featured in \cite{rakesh2023moving} to test object tracking with a Gaussian mixture model and in \cite{kumar2024handling} to test a method of background modeling under variations in illumination across the frames. We further comment that tests were also performed on the Background Models Challenge Database \cite{BMC}, but results were felt to be unsatisfying for presentation herein since there is little movement into and out of the videos to detect with our method. Nonetheless, results of background separation with DMD were presented previously in Figure~\ref{fig:background_foreground}. 

The Microsoft Wallflower dataset contains seven videos that are briefly described as follows. 
\begin{itemize}
    \item {\bf Camouflage (CAM)}: A scene of a computer monitor with interference bars; a man enters and exits the scene. 
    \item {\bf ForegroundAperture (FA)}: A person is sleeping at a desk. They arise and exit the frame, then return and remain motionless at the desk before leaving again. 
    \item {\bf LightSwitch (LS)}: A computer lab scene; a person enters the room, turns the light off and leaves; another person enters, turns on the light, and remains for a while before leaving. 
    \item {\bf MovedObject (MO)}: A conference room scene; a person enters and pulls up to a chair to use the telephone, leaves the scene, and returns to use the chair and phone again, leaving them in a different position each time. 
    \item {\bf TimeOfDay (TOD)}: A room with couches gradually illuminates from darkness, a person enters and exits twice to sit on the couch; the scene then gradually returns to darkness. 
    \item {\bf WavingTree (WT)}: An outdoor scene in daylight with a tree swaying in the background; a person briefly enters and exits the scene. 
    \item {\bf Bootstrap (BS)}: An overhead view of a busy cafeteria scene with many people entering and exiting the camera's view. Due to the presence of many subjects moving into and out of the frame constantly, our algorithm is unable to discern any foreground motion. 
\end{itemize}
We reiterate that, due to the nature of sliding window DMD, events in the first $T$ and last $T$ frames of the video are outside the scope of detection since there is nothing to compare the DMD spectrum with yet. Thus, any event(s) in the database videos that fall into this regime will not be incorporated into the error score calculation. Table~\ref{tab:microsoft_wallflower} provides both the optimal threshold value for each of the above videos and the corresponding error score. All other parameter values are fixed across all videos in this database and are the same as in the previous tests from our own curated database. 

\begin{table}[t]
    \centering
    \begin{tabularx}{0.9\textwidth}{l|XXXXXXX}
        \toprule
        Video name: & CAM & FA & LS & MO & TOD & WT & BS \\
        \midrule
        Error Score & 1 & 16 & 24 & 1 & 15 & 2 & N/A \\
        Optimal Threshold & 0.1530 & 0.5020 & 0.3850 & 0.5400 & 0.5010 & 0.1850 & N/A \\
        \bottomrule
    \end{tabularx}
    \caption{Result of applying the motion detection algorithm to the Microsoft Wallflower database.}
    \label{tab:microsoft_wallflower}
\end{table}

We can see from Table~\ref{tab:microsoft_wallflower} that our method performs well on three of the videos (CAM, MO, and WT), with only one or two false positives detected. The strong performance on WT and CAM in particular is an improvement over the results in \cite{rakesh2023moving}, where their method displayed some difficulties with these videos. Alternatively, we produce more false positives on the remaining three videos (FA, LS, and TOD). Two out of the three latter videos (LS and TOD) feature more pronounced lighting changes than the other videos, which could explain the algorithm's difficulty in detecting foreground motion since this is a well-known issue for many motion detection algorithms (see \cite{kumar2024handling} and the references therein). Overall, we can see that the optimal threshold varies widely for each video. This emphasizes the importance of recalibrating the algorithm for every new setting, as the same threshold value may not work well across varied settings. 



}

\section{Conclusion}\label{sec:Conclusion}

In this paper, we have outlined a new method to automatically detect motion in video data. Our method is firmly rooted in DMD, taking advantage of its linearity to detect motion through changes in the eigenvalues of the DMD matrix. While the method employs a number of parameters, the one that matters the most is the detection threshold. Thus, after demonstrating performance on individual videos, we turned to demonstrating the effect the threshold can have on the method and presented a framework in order to identify a threshold value that optimizes performance. 

The method relies on spikes in the DMD spectrum coming from sudden motion in the frame, typically from an individual entering or exiting it. {\color{black} The height of the spikes is likely to be related to the speed of movement entering the frame, since larger DMD eigenvalues are required to produce faster movement in the DMD decomposition using the temporal components $\mathrm{e}^{\omega_m t}$. } Thus, one possible way to fool this method of motion detection is to move extremely slowly so that eigenvalue spikes remain below the threshold. {\color{black} 
Furthermore, the spikes in the DMD spectrum related to movement are often short-lived (see Figure~\ref{fig:detection}) and so it is unclear if their duration indicates anything important about other movement factors such as volume and number of objects. Despite these potential shortcomings, our work herein has shown that DMD can be applied to motion detection to produce a real-time, flexible, and efficient method that can also be used to isolate foreground motion once it has been detected.  }

%

\section*{Availability of Data and Material}
{\color{black} All code to reproduce the demonstrations in this manuscript is available in the repository \href{https://github.com/marco-mig/dmd-motion-detection}{https://github.com/marco-mig/dmd-motion-detection}. The demonstrations in Section~\ref{sec:OtherVideos} use videos from the Microsoft Wallflower database, available at \href{https://www.microsoft.com/en-us/download/details.aspx?id=54651.}{https://www.microsoft.com/en-us/download/details.aspx?id=54651.} Our own video database will be made available to interested users upon reasonable request.}

\section*{Competing Interests}
The authors declare that they have no competing interests.

\section*{Funding}
This research was supported by an NSERC USRA (MM) and NSERC Discovery Grants (SB and JJB).

\section*{Authors' Contributions}
All authors contributed to the study conception and design. Numerical experiments were performed by MM, under the supervision of SB and JJB. All authors contributed equally to drafting the previous versions of the manuscript, as well as reading and approving the final manuscript. MM created all figures.

\section*{Acknowledgements}
The authors gratefully acknowledge Bianca Boissonneault, Lisa Marino and Miranda Mignacca for agreeing to be in our test videos.

\bibliographystyle{plain}
\bibliography{bibliography.bib}

\end{document}